\def\year{2020}\relax
\documentclass[letterpaper]{article} 
\usepackage{aaai20} 
\usepackage{times} 
\usepackage{helvet} 
\usepackage{courier} 
\usepackage[hyphens]{url} 
\usepackage{graphicx} 
\usepackage{graphicx} 
\usepackage{tikz}
\usetikzlibrary{shapes,arrows,positioning}

\title{Face-work for Human-Agent Joint Decision-Making}

\author{\Large \textbf{JiHyun Jeong}\\
Information Science \\
Cornell University, Ithaca, NY\\
jihyun@infosci.cornell.edu
\And
\Large \textbf{Guy Hoffman}\\
Mechanical and Aerospace Engineering\\
Cornell University, Ithaca, NY\\
hoffman@cornell.edu
}

\begin{document}
 \newcommand{\citet}[1]{\citeauthor{#1} \shortcite{#1}} \newcommand{\citep}{\cite}
 \newcommand{\citealp}[1]{\citeauthor{#1} \citeyear{#1}}
\maketitle

\begin{abstract}
   We propose a method to integrate face-work, a common social ritual related to trust, into a decision-making agent that works collaboratively with a human. 
   Face-work is a set of trust-building behaviors designed to ``save face'' or prevent others from ``losing face.''
   This paper describes the design of a decision-making process that explicitly considers face-work as part of its action selection. We also present a simulated robot arm deployed in an online environment that can be used to evaluate the proposed method. 
\end{abstract}

\section{Introduction}
This paper describes a human-robot joint decision-making algorithm that integrates the social ritual of \emph{face-work}. Face-work involves the actions taken to maintain one's own or another's ``face'', a positive self-image claimed in a social context~\citep{goffman1967face}. We first formalize a synchronous decision-making interaction between a human and a robot and present a simulation setup to elicit such interactions. We then illustrate the implementation of an algorithm that evaluates decisions while accounting for face threats in the action-selection process. 

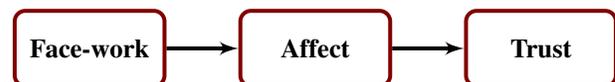
\begin{figure}[b]
    \centering
    \tikzstyle{block} = [rectangle, draw=red!50!black, font=\bfseries, text width=5em, text centered, on grid, line width=1.5pt, rounded corners, minimum height=1cm]
    \tikzstyle{arrowline} = [draw, line width=1.5pt, -latex']
    \begin{tikzpicture}[auto]
        \node [block] at (-3,0) (facework) {Face-work};
        \node [block] at (0,0) (affect) {Affect};
        \node [block] at (3,0) (trust) {Trust};
        \path [arrowline] (facework.east) to (affect.west);
        \path [arrowline] (affect.east) to (trust.west);
    \end{tikzpicture}
    \caption{Humans perform face-work to protect emotions attached to face~\cite{goffman1967face}. Affect-based elements such as care and concern for another constituted a core basis for trust in peer relationships~\cite{mcallister1995affect}.}
    \label{fig:diagram}
\end{figure}

When people collaborate, especially on making decisions, the affect-based trust they have in each other is crucial to make good decisions together. The importance of affective trust is demonstrated in multiple studies; for a review, see~\citet{lee2004trust}. In particular, care and concern for another constitute a core basis for trust in peer relationships~\cite{mcallister1995affect}. Figure~\ref{fig:diagram} illustrates this relationship between face-work, affect, and trust. 

To maintain affective trust, people engage in social rituals that are designed to ``save face'' for themselves, or prevent others from ``losing face''. The loss of one's face, which can be induced by criticism or disagreement, is often accompanied by negative emotions such as shame~\citep{goffman1967face}. For example, if a collaborator publicly dismisses a peer, the loss of face of the peer can prevent them from raising future ideas. Given the importance of face in affective trust, humans commit to face-work as a social ritual to defend against and correct face threats.

Similar social and interaction aspects are also pertinent when humans and agents make joint decisions. Achieving productive decision-making with humans and algorithms can be challenging~\citep{Green2019} and immediate performance gain is not always the best goal to strive for. 
For example, \citet{Elmalech2015} present cases when an algorithm that suggests a suboptimal option better matching a human's intuition leads to better performance over time, compared to suggesting optimal solutions that counter human intuitions. We posit that face-related factors can be another motivation for an algorithm or robot to prefer a suboptimal suggestion, i.e., when the optimal decision is a face-threat that can hamper social relationships.

Prior work in human-computer interaction already suggests that an agent can perform face-work to prevent humans from losing face, which can result in negative emotions that undermine affect-based trust.
~\citet{reeves1996media} found that humans treat computers as social actors and can take offense to impolite computers. Similarly, \citet{takayama2009m} found that human subjects were sensitive to disagreements from robots. \citet{jung2017affective} emphasize the need for affective grounding in human-robot collaborations. In our prior work, humans ascribed social intentions to robot actions, and a robot's misaligned action was sometimes interpreted as contemptuous and resulted in mistrust~\cite{Law2019}.

\begin{figure*}[ht]
    \centering
    \includegraphics[width=0.9\textwidth]{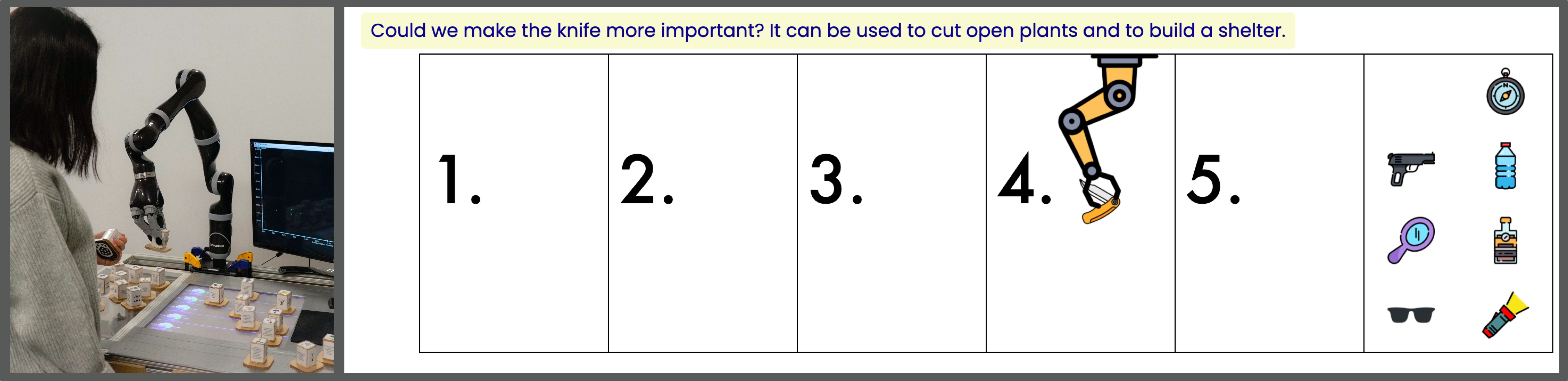}
    \caption{The user interface on the right allows both the human and the virtual robot to move objects to a box that corresponds to the preferred rank of the object. The interaction is designed to resemble a human-robot decision-making setup we used in prior work (left). Individual icons are from www.flaticon.com, and a list of credited authors can be found on our online platform.}
    \label{fig:sys}
\end{figure*}

Politeness is one type of face-saving strategy, and in fact has been studied extensively in human-agent and human-robot interaction. Research has found that applying politeness strategies improved perception of a robot and increased its trustworthiness. Participants felt more positive towards a robot that applied a distancing politeness strategy, having the disagreeing voice come from a box separate from the robot's body~\citep{takayama2009m}. Robotic assistants that gave advice using hedges and polite discourse markers were perceived as more considerate and likeable, and less controlling~\cite{torrey2013robot}. Similarly, politeness improved perceptions such as fairness and friendliness of a access-control robot~\cite{inbar2019politeness}.
Agents that politely engaged in small talk were viewed as more trustworthy by extroverts~\cite{bickmore2001relational}, while automation that had ``poor etiquette'' of interrupting humans, were trusted less~\cite{parasuraman2004trust,miller2005trust}.

Our work extends this prior literature in two ways. 
First, the robot or agent in the above studies was almost always in an assistive role, and the human had the sole agency to accept or reject the robot's suggestions. In contrast, we consider an interaction that resembles a process between equal partners with shared agency. This means that both human and the robot can accept or reject each other's suggestions. They continually negotiate their ideas and preferences back and forth, which provides opportunities for social breakdowns.

Second, prior works study politeness as a possible predictive construct in human-robot and human-agent interaction.
In our work, we propose to systematically include face-work in the decision-making process of the robot. 
We do so by presenting a computational framework for joint decision-making that integrates face-work in the agent's algorithm. We also describe a simulated environment that can be used to evaluate this framework.

\section{Joint Decision-Making Context}
To demonstrate the integration of face-work in human-robot interaction, we implement a simulator with a virtual robot. We chose this method for ease of user testing while keeping some physical affordances of robots (Figure~\ref{fig:sys}). This way we can use spatial cues as embodied elements (e.g., for epistemic actions) and use the robot's nonverbal behavior for face-saving gestures. 

\subsection{Adapted Desert Survival Task}
Our task is an adapted version of the Desert Survival Problem~\cite{lafferty1974desert}. A participant first ranks a set of given items in the order of importance for survival, then collaborates with a teammate to arrive at a group solution. The problem and its variations have been used widely in teamwork and group decision-making studies~\cite{Burke2015,Hall1970}.

We adapt the task to be applicable to more general group decisions where a group not only rank, but also accept and reject different options. Instead of having the participants rank all items, we ask them to choose five out of eight items and rank only those they have selected. Selecting the top five presents a clear bound between accepted and rejected items, which is the case in many group decisions. For instance, when a group engages in brainstorming, some ideas will be accepted, some rejected, and some may be preferred over others.

\subsection{Interface and Procedure}

Decisions to rank an item are expressed by moving an object to a location that corresponds to the desired rank. Both parties can add an object to any rank, remove an object, or swap the location of two objects. The human ranks an object by dragging the icon representing the object to a box with the preferred rank. The virtually embodied robot moves an object through a series of animations: moving to the location of an object, picking it up, moving to the designated rank, and dropping the object. The series of animations take around 7 seconds to move one object, which is slower than the speed at which a human is capable of dragging and dropping an object. This design preserves the difference in speed between a human and a real-world collaborative robot.

The human and the robot take turns to negotiate the ranking. 
To simulate a real-world interaction more accurately, we implemented a flexible turn-taking paradigm: 
The human can choose to move as many objects as they want for a prolonged period of time in one turn. During that time, the robot does not move any objects. When a human wants to yield their floor to the robot, they can pause for the robot to start moving. The robot will then move objects until the human takes the floor back, which can only happen during a robot pause. Manipulating the length of robot pause times allows for the robot to also take multiple actions in one turn. 

In each move, the robot provides a reason for ranking an object higher or lower. The final team rank can be submitted when the human and the robot both agree on a solution.

\section{Agent Design}
We adapt a formalization of the survival task from~\citet{Bergner2016}. Each object has an integer identifier $i$, and a ranking at time step $j$ is represented as an array $R_j$ of eight numbers corresponding to the location (rank) of each object. Different objects can share the rank $6$ when they are not ranked in the top five. A robot move at turn $j$ is represented by a tuple $m_j = (i, orig, dest)$ including the identifier of the object in the array, its current (origin) rank and its destination rank. 

The robot considers each possible move candidate $m_j$ and the ranking it produces, $R_j$, by applying $m_j$ to its previous state of ranking $R_{j-1}$. It then evaluates how far each candidate ranking is to the robot's preferred ranking. We use the sum of two widely used evaluation metrics for this task, minimizing distance ($D(R_j)$) from the current ranking to the desired ranking~\cite{burgoon2000interactivity} and maximizing the number of concordant pairs ($C(R_j)$) between the two~\cite{Bergner2016}. 

\subsection{Integrating Machine Face-Work}
We define an additional measure of decorum based on face-work, $F(m_j)$, for a move $m_j$. Thus, each move $m_j$ that achieves ranking $R_j$, will be evaluated using a combination of metric that evaluates the ranking produced by a move, together with an assessment of how socially appropriate the move is. The robot will execute the action $m_j$ that leads to the maximum value of $ (C(R_j)-D(R_j)) * F(m_j)$.

To determine a valid model for $F$, we incorporate the following politeness strategies (adapted from \citet{brown1987politeness}) to mitigate face threatening acts: seek agreement, avoid disagreements, and make indirect requests. First, for all candidate moves $m_j$, $F$ will be initialized to $F(m_j) = 1$. Next, we illustrate how we compute $F(m_j)$ for each candidate.

\subsubsection{Seek agreement:}
The robot looks for items that it and the human both agree on to rank first. Let us represent the robot's preferred ranking\footnote{We use $A$ for ``agent'' to avoid confusion with the notation for ranking} as $R^A=[r_1^A,\ldots,r_8^A]$ and the human's preferred ranking, expressed in the initial stage of the task when working on their own, as $R^H = [r_1^H, \ldots,r_{8}^H]$. For any move $m_j = (i, orig, dest)$ where $dest = r_i^A = r_i^H$, has the value of $F(m_j)=2$.

\subsubsection{Avoid disagreements:}
We avoid any moves that directly reverse any actions that the human has done in their previous turn. A human's last turn can be represented as an array consisting of tuples of human moves. For convenience, we represent all human moves on their last turn $m_{H_1},m_{H_2},\ldots,m_{H_n}$ as $m_H = (i_H,orig_H, dest_H)$. Any candidate robot move $m_j = (i, orig, dest)$ that meets $i= i_H$ and $dest = orig_H$ is considered a reversal, and will have the value of $F(m_j)=0$.

We also avoid any moves that repeat any of the robot's actions, since the need to do so implies that the human had disagreed with the choice before and might be viewed as a dismissing insistence. We represent all previous robot moves $m_{A_1},m_{A_2},\ldots,m_{A_k}$, as $m_A = (i_A,orig_A, dest_A)$. Any candidate robot move $m_j = (i, orig, dest)$ that meets $i = i_A$ and $dest = dest_A$ is considered a repeat, and will have the value of $F(m_j)=0$.

A useful side-effect of using the above face-work evaluation function is that it can also implicitly signal an ostensible compromise. For instance, if the human places an item on the third rank after the robot placed it on first, the robot will then try placing it on second instead to avoid reversing the human's previous decision.

\subsubsection{Make indirect requests:}
Lastly, we also incorporate verbal politeness markers and make indirect requests using questions and hedges. For example, the robot asks ``Could we make the knife more important?'' instead of saying ``Make the knife more important.''

\section{Evaluation Plan}
 We plan to conduct experiments to test whether these strategies are effective at maintaining human's trust. As a first step, we designed a within-subjects study, where participants interact with both agents, one that performs face-work as indicated above and one that does not. The order in which these agents are presented is randomized. There are two variations of the survival task, for which the order is also randomized. This measure is taken to avoid preferences carrying over between conditions. We will perform a manipulation check to confirm that our agent can reliably mitigate face-threats using scales such as the Revised Instructional Face-Support (RIFS) scale~\cite{KerssenGriep2008}. Lastly, human's trust towards the agent will be measured using validated questionnaires such as those proposed by \citet{madsen2000measuring}.

\section{Conclusion}
This work is part of an on-going work of developing socially intelligent agents that can maintain human's trust in group interactions. We introduce a joint decision-making setting between a human and a virtual agent, and implement face-work into the algorithm of the agent's decision.

In future work, this platform can host more intelligent agents to interact with humans that considers more of the social contexts and relationships. We make a crass distinction of face-threats in this work and only deal with one mitigation strategy: to avoid it. However, there are degrees of impositions and a spectrum of mitigation strategies. We can also incorporate different social factors from literature that might determine the acceptable degree of face threats, such as power or social distance. Additionally, future work will also examine ethical challenges of developing a socially intelligent agent that learns to negotiate, as there can be negative implications for socially deceptive behaviors that can manipulate a person's affect and trust.

\section{Acknowledgements}
This material is based upon work supported by the Defense Advanced Research Projects Agency (DARPA) under Contract No. W911NF2010004. Any opinions, findings and conclusions or recommendations expressed in this material are those of the authors and do not necessarily reflect the views of the Defense Advanced Research Projects Agency (DARPA).

\bibliography{jeong-aaai-fss20}
\bibliographystyle{aaai}

\end{document}